\documentclass[conference]{IEEEtran}
\IEEEoverridecommandlockouts
\usepackage{cite}
\usepackage{amsmath,amssymb,amsfonts}
\usepackage{caption}
\usepackage{graphicx}
\usepackage{textcomp}
\usepackage{xcolor}
\usepackage{multirow}
\usepackage{subcaption}
\usepackage{algorithm}
\usepackage[noend]{algpseudocode}
\usepackage{hyperref}       
\usepackage{url}            
\def\BibTeX{{\rm B\kern-.05em{\sc i\kern-.025em b}\kern-.08em
    T\kern-.1667em\lower.7ex\hbox{E}\kern-.125emX}}

\DeclareMathOperator*{\argmin}{arg\,min}
    
\begin{document}

\title{HINT: Healthy Influential-Noise based Training to Defend against Data Poisoning Attacks}

\author{
\IEEEauthorblockN{Minh-Hao Van, Alycia N. Carey, Xintao Wu}
\IEEEauthorblockA{
\textit{University of Arkansas}\\
Fayetteville, AR, USA \\
\{haovan, ancarey, xintaowu\}@uark.edu}
}

\maketitle

\begin{abstract}
While numerous defense methods have been proposed to prohibit potential poisoning attacks from untrusted data sources, most research works only defend against specific attacks, which leaves many avenues for an adversary to exploit. In this work, we propose an efficient and robust training approach to defend against data poisoning attacks based on influence functions, named \textit{Healthy Influential-Noise based Training}. Using influence functions, we craft healthy noise that helps to harden the classification model against poisoning attacks without significantly affecting the generalization ability on test data. In addition, our method can perform effectively when only a subset of the training data is modified, instead of the current method of adding noise to all examples that has been used in several previous works. We conduct comprehensive evaluations over two image datasets with state-of-the-art poisoning attacks under different realistic attack scenarios. Our empirical results show that HINT can efficiently protect deep learning models against the effect of both untargeted and targeted poisoning attacks.
\end{abstract}


\section{Introduction}
Having access to high-quality, clean, and human-annotated data is essential to building and training a well-performing prediction model. However, it is common that an organization only has a limited amount of this type of data on hand. Consequently, organizations are often tasked with collecting additional data from outside sources using techniques such as web scraping and/or crowd-sourcing. This, unfortunately, opens numerous avenues for attacking the proposed model, such as \textit{data poisoning} attacks in which an attacker injects harmful data into the training routine to affect the final model's utility. For example, in \cite{DBLP:conf/nips/FengCZ19,DBLP:conf/iclr/MadryMSTV18}, the authors demonstrate the ability of adversarially crafted examples to destroy a DNN's prediction accuracy and \cite{DBLP:conf/eurosp/AghakhaniMWKV21, DBLP:conf/iclr/GeipingFHCT0G21, DBLP:conf/nips/HuangGFTG20,DBLP:conf/nips/ShafahiHNSSDG18} show that attackers can force a model to predict the adversarial class on a specific targeted example by only having to modify a small fraction of the training data.

Telling if a certain collected data point is benign or malicious is a non-na\"{i}ve task, and there is an active area of research focused on building defense methods against poisoning attacks as well as analyzing the harm that poisoning attacks have on the final model\cite{geiping2021doesn, liu2022friendly, DBLP:conf/nips/TaoFYHC21, DBLP:conf/icml/YangLM22}. From a defense perspective, most of the proposed works are attack-specific and are easily defeated by newer types of attacks that consider the underlying defense mechanism. Some other defenses focus on pre-processing the training data in order to detect and remove malicious examples before they are used for training \cite{DBLP:conf/bigdataconf/VanDWCL22,DBLP:conf/aaai/ChenCBLELMS19,DBLP:conf/iciot/BaracaldoCLSZ18,DBLP:conf/nips/Tran0M18}. The pre-processing approach works well when the malicious perturbations are large, or when only a fraction of the dataset is poisoned. However, when those criteria are not met, the defenses based on pre-processing are easily overcome \cite{geiping2021doesn,liu2022friendly}. Another problem inherent to current research on defending against poisoning attacks is the trade-off between model accuracy and the effectiveness of a defense -- especially in cases where DNN's are used as the model architecture. Although some modern defense mechanisms have the ability to achieve better generalization \cite{liu2022friendly,DBLP:conf/icml/YangLM22,DBLP:conf/bigdataconf/VanDWCL22}, the need for more research remains. 

In this work, we consider the realistic scenario of the training dataset only containing a limited number of clean and human-labeled data points, while the remaining points are unverified and malicious data collected from untrustworthy outside sources. In order to defend a classifier against poisoning attacks, we propose Healthy Influential-Noise based Training (HINT)\footnote{We interchangeably use \textit{healthy noise} as an alternative to \textit{healthy influential-noise} in the paper} -- a robust training procedure based on influence functions. Influence functions, originally a product of robust statistics \cite{cook1982residuals}, have grown popular over the last few years as an explainability method for understanding black-box model predictions \cite{DBLP:conf/icml/KohL17}. In this work, we show that influence functions, in addition to explaining the effect an entire training point has on the model parameters and/or test loss, can capture useful information about the impact of each local pixel to the model's prediction. 
Consequently, those pixels, which are identified as influential, form local regions that cause significant changes in the test loss of the model. By incorporating HINT in the training procedure, we are able to: (1) identify a subset of training examples that have high impact on the model loss, (2) craft the healthy influential-noises that both reduce the harmful regions and boost the helpful regions inside images, and then add them into training examples to reduce the effect of poisoning attacks. HINT can help the trained model predict the correct class with high confidence score. Through extensive experiments, we show that the classification model trained with our HINT can resist different types of untargeted and targeted attacks while retaining good generalization.

The remainder of the paper is as follows. In Section \ref{sec:related-work}, we detail closely related works in poisoning attacks and defenses. In Section \ref{sec:prelims}, we introduce the influence function, the central mechanism in our method. Section \ref{sec:hint} details our HINT method. Section \ref{sec:atk_def} discusses data poisoning attacks and defenses that we use in our evaluation. Then, Section \ref{sec:experiments} shows our experiments on HINT and other baselines. Finally, we offer our concluding remarks in Section \ref{sec:conclusion}.

\section{Related Work}
\label{sec:related-work}
\subsection{Poisoning Attacks}
Poisoning attacks, which manipulate the training data to compromise the model's performance at test time, can be grouped into two categories: \textit{untargeted attacks} (or availability attacks) and \textit{targeted attacks} (or integrity attacks). In untargeted attacks, the attacker manipulates a subset of the training data to degrade the utility of the machine learning model in general, and are originally proposed to attack traditional classification models such as linear regression and support vector machines \cite{DBLP:journals/jmlr/BiggioNL11,DBLP:conf/icml/BiggioNL12,DBLP:conf/ccs/NewellPXN14,DBLP:conf/ijcai/ZhaoAGZ17}. In the deep learning setting, there are few modern untargeted attacks \cite{DBLP:conf/nips/FengCZ19,DBLP:conf/iclr/MadryMSTV18,DBLP:conf/ccs/Munoz-GonzalezB17} that focus on threatening model availability. In addition to performing attacks on DNN models, untargeted attacks focusing on affecting the fairness of a model have been proposed. Specifically, \cite{chang2020adversarial, DBLP:conf/dasfaa/VanDWL22} both proposed untargeted poisoning attacks on fair machine learning models and demonstrated the trade-off between fairness and accuracy. 

Contrasting untargeted attacks, instead of attempting to sabotage the model in general, targeted attacks aim to undermine the integrity of a specific test example (or a set of test examples), which are more challenging to defend against than untargeted attacks. The victim model trained on poisoned data crafted using targeted attacks still achieves good overall accuracy, but the predictions on the targeted examples (selected by the attacker) are misclassified into the intended adversarial class. Many proposed attacks \cite{DBLP:conf/eurosp/AghakhaniMWKV21, DBLP:conf/iclr/GeipingFHCT0G21, DBLP:conf/nips/HuangGFTG20,DBLP:conf/nips/ShafahiHNSSDG18} can successfully cause a deep learning model (e.g., ResNet or VGG) to predict, with high probability, the adversarial class for a targeted image instead of the actual class. 

\subsection{Defenses Against Poisoning Attacks}
There are two main strategies to defend against data poisoning attacks: \textit{filtering defense} and \textit{robust training}. \textit{Filtering defense} aims to detect malicious examples in the training data and intervene before they can harm the model. The most common filtering defenses are: 1) applying pre-processing techniques on a pre-trained model; and 2) implementing in-processing strategies during the training phase. In \cite{DBLP:conf/aaai/ChenCBLELMS19}, the authors used clustering methods on the activation layers of a neural network to detect poisons in the training set. 
Data provenance is used in \cite{DBLP:conf/iciot/BaracaldoCLSZ18} to identify poisoned data by evaluating the likelihood of a data point being poisoned. \cite{DBLP:conf/nips/Tran0M18} showed that strong signals in hidden representations often mean that a data point has been attacked. Their method, in turn, examines the distribution shift between malicious and clean inputs to detect and remove poisoned examples. \cite{DBLP:conf/icml/YangLM22} proposed EPIC, an effective defense that performs filtering during the training phase. 
The common assumption of all these attacks is that the overall fraction of malicious examples in the training set is small, and removing them does not hurt the model's generalization ability. Moreover, heavy computational resources are required to choose the optimal filtering settings for each method correctly. In contrast, our HINT method does not aim to remove malicious examples from the training data. Instead, we generate healthy noise such that when added to an image, it alleviates the effect of the poisoned data. 

\textit{Robust training} methods usually apply smoothing and augmentation techniques to make the model more robust to noisy data. In \cite{DBLP:conf/sp/WeberXKZL23}, the authors introduced a unified framework to deal with poisoning attacks via randomized smoothing. From the augmentation approach, \cite{DBLP:conf/icassp/BorgniaCFGGGGG21} proposed to use strong data augmentation such as MixUp while \cite{borgnia2021dp} combined MixUp with random smoothing noise from differentially private training to achieve more robust defense. \cite{geiping2021doesn} and  \cite{DBLP:conf/nips/TaoFYHC21} both leveraged the idea of adversarial training, which was proposed initially to deal with evasion attacks, to defend against poisoning attacks. While 
\cite{DBLP:conf/nips/TaoFYHC21} aims to perform adversarial training against delusive attacks (a.k.a clean-label availability attacks), \cite{geiping2021doesn} simulated the attacks during the training phase by creating and injecting targeted poisoning attacks into training data. In \cite{liu2022friendly}, the authors proposed optimizing two components, friendly noise, and random noise, to perturb training examples so that they can alleviate the harmful effects of poisoned data without losing the generalization ability of the model. Differentially private SGD (DP-SGD) has also been proposed as a strategy to train a robust model against poisoning attacks \cite{DBLP:conf/ccs/AbadiCGMMT016,DBLP:conf/uss/Jayaraman019}. 

\section{Preliminaries}
\label{sec:prelims}
Inspired by the influence function from robust statistics \cite{cook1982residuals}, Koh and Liang \cite{DBLP:conf/icml/KohL17} introduced a method for estimating the influence that a training point $z=(x,y)$ has on a machine learning model, where $x \in X$ in the input and $y \in Y$ is the class. Let $f_\theta$ be a classification model parameterized by $\theta$ and let $D_{trn}/D_{val}/D_{tst}$ be the training/validation/test sets. Let $l(\cdot, \theta)$ represent the loss and $L(D_{trn},\theta) = \frac{1}{|D_{trn}|}\sum_{z_i\in D_{trn}}l(z_i, \theta)$ be the empirical loss to be minimized during training. To see the change in model parameters w.r.t to training point $z$, the ERM formulation can be modified as:
\begin{equation}
    \label{eq:erm}
    \hat{\theta}_{\epsilon, z}=\argmin_{\theta \in \Theta}\dfrac{1}{|D_{trn}|}\sum_{z_i \in D_{trn}}l(z_i,\theta)+\epsilon l(z, \theta)
\end{equation}
where $z$ is effectively upweighted by a small weight $\epsilon$ (usually on order of $\frac{1}{n}$ where $n$ is the number of training points). Instead of actually performing training using Eq. \ref{eq:erm}, \cite{DBLP:conf/icml/KohL17} shows that it can be estimated without actually having to retrain the model on $D_{trn}\setminus\{z\}$:
\begin{equation}
\label{eq:up-param}
    \mathcal{I}_{up, param}(z) = \dfrac{d\hat{\theta}_{\epsilon,z}}{d\epsilon}\Big|_{\epsilon=0}=-H^{-1}_{\hat{\theta}}\nabla_{\theta}l(z,\hat{\theta})
\end{equation}
In addition to showing the effect a training point $z$ has on the parameters, Eq. \ref{eq:up-param} can be extended to show the influence that $z$ has on a test point $z_{test}$.
\begin{equation}
    \mathcal{I}_{up, loss}(z, z_{test}) = -\nabla_{\theta}l(z_{test}, \hat{\theta})^{\top}H_{\hat{\theta}}^{-1}\nabla_{\theta}l(z, \hat{\theta}) 
    \label{eq:ss_up_loss}
\end{equation}
Eq. \ref{eq:up-param} and Eq. \ref{eq:ss_up_loss} simulate the effect of $z$ being removed from the dataset. However, the effect of perturbing $z$ can be estimated via influence functions as well. Let $\hat{z} = (x+\delta,y)$ be the perturbed variant of $z$ by adding a small noise $\delta$. One can define the parameters resulting from moving $\epsilon$ mass from $z$ onto $\hat{z}$ as:
\begin{equation}
    \label{ref:approx-param}
    \hat{\theta}_{\epsilon, \hat{z}, -z} = \argmin_{\theta \in \Theta} \dfrac{1}{|D_{trn}|}\sum_{z_i\in D_{trn}}l(z_i, \theta)+\epsilon l(\hat{z}, \theta)-\epsilon l(z,\theta)
\end{equation}
Analogous to Eq. \ref{ref:approx-param}, Eq. \ref{eq:pert-param} uses the influence function to approximate the effect that modifying a training point $z\to\hat{z}$ has on the model parameters:
\begin{equation}
\label{eq:pert-param}
    \begin{aligned}
    \mathcal{I}_{pert,param}(z) &=  \dfrac{d\hat{\theta}_{\epsilon,\hat{z},-z}}{d\epsilon}\Big|_{\epsilon=0} \\
     &= \mathcal{I}_{up, param}(\hat{z})-\mathcal{I}_{up, param}(z) \\
    &= -H^{-1}_{\hat{\theta}}\left(\nabla_\theta l(\hat{z}, \hat{\theta})-\nabla_\theta l(z, \hat{\theta})\right)
\end{aligned}
\end{equation}
As in the case with Eq. \ref{eq:up-param}, \cite{DBLP:conf/icml/KohL17} extends Eq. \ref{eq:pert-param} to show how perturbing $z\to \hat{z}$ would affect the loss of a test point $z_{test}$:
\begin{equation}
    \mathcal{I}_{pert, loss}(z, z_{test}) = -\nabla_{\theta}l(z_{test}, \hat{\theta})^{\top}H_{\hat{\theta}}^{-1}\nabla_x\nabla_{\theta}l(z, \hat{\theta})
    \label{eq:ss_pert_loss}
\end{equation}
The main difference between Eq. \ref{eq:ss_up_loss} and Eq. \ref{eq:ss_pert_loss} is that in Eq. \ref{eq:ss_pert_loss}, the gradient of $\nabla_{\theta}L(z, \hat{\theta})$ w.r.t $x$ is additionally calculated. This additional gradient computation captures how changing $z$ along each dimension of $x$ affects the loss of a test point. 


\section{Healthy Influential-Noise based Training}
\label{sec:hint}
In this section, we propose Healthy Influential-Noise based Training (HINT) which is a training procedure robust to malicious training examples. Our HINT reduces the potential harm caused by untrusted data sources while retaining the model's generalization ability.

\subsection{Framework}
\label{sec:framework}
Most robust training approaches manipulate the training data (or a subset of the training data) to train a robust model. In our method, we construct a subset $D_s\subset D_{trn}$ by choosing the most influential examples in $D_{trn}$. Additionally, we denote the training points not chosen to be in the subset as $D_u = D_{trn}\setminus D_s$. Let $\delta_i$ is the healthy influential-noise and $\hat{z}_i=(x_i+\delta_i, y_i)$ be the healthy-perturbed version of $z_i = (x_i, y_i)$. With image data, $\delta_i$ lies in the space $\Delta = \{\delta \in \mathbb{R}^{H\times W}: \|\delta\|_{\infty} \leq \beta\}$, where $\|\cdot\|_\infty$ is the $L_\infty$-norm and $\beta$ is the hyper-parameter for bounding the noise. 

We define the healthy-perturbed training set as $\hat{D}_{trn}=\hat{D}_s\cup D_u$, where $\hat{D}_s$ is the set $D_s$ \textit{after} healthy noise is added. Under this setting, we define the empirical loss function of the defense model as $L(\hat{D}_{trn}, \theta) = L(\hat{D}_{s}, \theta) + L(D_u, \theta)$, where $L(D_{\ast}, \theta) = \dfrac{1}{|D_{\ast}|}\sum_{z_i \in D_{\ast}}l(z_i, \theta)$ and $D_{\ast}$ denotes ``any dataset''. We define the defender's objective as:
\begin{equation}
    \min_{\delta \in \Delta} \ L(D_{val}, \theta_{\delta}) \text{ s.t. } \theta_{\delta} = \argmin_{\theta \in \Theta} L(\hat{D}_{trn}, \theta) \label{eq:noise_train}
\end{equation}
The na\"{i}ve approach to the above problem would be for the defender to try several different healthy noise values $\delta_i$ and to train/optimize several different models. However, this na\"{i}ve approach is intractable since the feasible spaces for $\Delta$ and $\Theta$ are sufficiently large, and it will take significant computational resources to train multiple instances of only one particular model architecture. By using the influence function, we avoid the requirement of costly retraining. Specifically, we use influence function to estimate the change in the model's loss on $D_{val}$ when modifying a particular training data point to efficiently defend against poisoning attacks.

\begin{algorithm}[t!]
    \begin{algorithmic}[1]
        \renewcommand{\algorithmicrequire}{ \textbf{Input:}}
        \renewcommand{\algorithmicensure}{ \textbf{Output:}}
        \Require Training data $D_{trn}$, validation data $D_{val}$, train epochs $T$, pre-train epochs $T_{pre}$, scaling factor $\gamma$, healthy noise bound $\beta$, ratio of selected examples $r$, learning rate $\eta$, healthy noise update schedule $S$
        \Ensure Trained model $\hat{\theta}$
        \State Initialize $\theta^0$
        \For{$t=1\dots T_{pre}$}
            \State $\theta^t \leftarrow \theta^{t} - \eta\nabla L(D_{trn},\theta^t) $
        \EndFor
        \State $D_s, D_u \leftarrow$ SecInf($D_{trn}$, $D_{val}$, $r$) using Algorithm \ref{alg:selection}
        \State $\hat{D}_{trn} \leftarrow D_{trn}$,  $\hat{D}_s \leftarrow D_s$ 
        \For{$t=T_{pre}+1\dots T$}
            \If{$t \in S$}
                \State $\hat{D}_s \leftarrow \text{AddNoise}(\hat{D}_s, D_s, \gamma, \beta)$ using Algorithm \ref{alg:noise_calc}
                \State $\hat{D}_{trn} \leftarrow \hat{D}_{s} \cup D_u$  
            \EndIf   
            \State $\theta^t \leftarrow \theta^{t} - \eta \nabla L(\hat{D}_{trn},\theta^t) $
        \EndFor
    \end{algorithmic}
    \caption{HINT: Healthy Influential-Noise based Training}
    \label{alg:hint}
\end{algorithm}

In this work, we introduce HINT, a training algorithm with healthy influential-noise, that: (1) selects a subset of training points $D_s$ which have the most impact on the model by calculating the influence of each training point on the model loss; and (2) generates healthy noise for every example in $D_s$ to reduce the success of poisoning attacks without significantly degrading the model performance over the test data. Algorithm \ref{alg:hint} fully presents HINT. First, in lines 2-3, we perform pre-training of the model parameterized by $\theta$ for a few epochs. This pre-training is to warm-up the $\theta$ parameter to avoid instability in early epochs. Lines 4-10 contain the main routine of HINT. In line 4, we  select the most influential training points $D_s$ using Algorithm \ref{alg:selection} (discussed in Section \ref{sec:secinf}). In lines 6-10, we first check if the round is an update round (which is defined by an update schedule $S$), and if it is, we generate and add healthy noise to selected training examples following Algorithm \ref{alg:noise_calc} (line 8), and then update $\hat{D}_{trn}$ (line 9). Details of generating healthy influential-noise will be given in Section \ref{sec:add_noise}. Regardless if the noise is updated or not, the model parameters are updated on $\hat{D}_{trn}$ in every epoch (line 10).

\noindent\textbf{Influence Function on Validation Group.} It is essential to note that Eqs. \ref{eq:ss_up_loss} and \ref{eq:ss_pert_loss} consider the influence that a single training point has on a single test point. Calculating the influence score with respect to only one single test point, however, may not produce a good estimation when the training data is poisoned. Additionally, it is computationally expensive to calculate the influence score for each pair of training and test points individually, since each pair requires the inverse Hessian matrix to be calculated (or estimated). Therefore, we extend the influence functions of Eqs. \ref{eq:ss_up_loss} and \ref{eq:ss_pert_loss} to estimate the impact that a single training point has on a group of test (or validation) points. Since influence is additive \cite{DBLP:conf/icml/KohL17}, we can extend both equations to consider the loss on a group of validation points. The influence of a training point on the loss of a validation set $D_{val}$, in both the total removal and perturbation cases are:
\begin{equation}\small
    \begin{split}
        \mathcal{I}_{up, loss}(z, D_{val}) = -\nabla_{\theta}L\left(D_{val}, \hat{\theta}\right)^{\top}H_{\hat{\theta}}^{-1}\nabla_{\theta}l\left(z, \hat{\theta}\right) \\
        = -\left[\nabla_{\theta}\dfrac{1}{|D_{val}|}\sum_{i=1}^{\left|D_{val}\right|} l\left(z_{i}, \hat{\theta}\right)\right]^{\top}H_{\hat{\theta}}^{-1}\nabla_{\theta}l\left(z, \hat{\theta}\right)\\
    \end{split}
    \label{eq:sg_up_loss}
\end{equation}

\begin{equation}\small
    \begin{split}
        \mathcal{I}_{pert, loss}(z, D_{val}) = -\nabla_{\theta}L\left(D_{val}, \hat{\theta}\right)^{\top}H_{\hat{\theta}}^{-1}\nabla_x\nabla_{\theta}l\left(z, \hat{\theta}\right)\\
        = -\left[\nabla_{\theta}\dfrac{1}{|D_{val}|}\sum_{i=1}^{\left|D_{val}\right|} l\left(z_{i}, \hat{\theta}\right)\right]^{\top}H_{\hat{\theta}}^{-1}\nabla_x\nabla_{\theta}l\left(z, \hat{\theta}\right)\\
    \end{split}
    \label{eq:sg_pert_loss}
\end{equation}
Note that for deep learning models, we can compute the influence score using only the top layers instead of the full network, which is a common way since the top layers work as a classifier and the bottom layers work as a feature extractor. Even if we only consider the top layers, computing the inverse Hessian matrix ($H_{\hat{\theta}}^{-1}$) is computationally intensive. To avoid the direct computation of the inverse Hessian matrix, we can instead leverage the inverse Hessian-Vector Product (IHVP) method to approximate $H_{\theta}^{-1}\nabla_{\theta}L(D_{val}, \hat{\theta})$ in Eq. \ref{eq:sg_up_loss} and Eq. \ref{eq:sg_pert_loss}. We use Linear time Stochastic Second-Order Algorithm (LiSSA) \cite{DBLP:journals/jmlr/AgarwalBH17} to compute the IHVP efficiently.

\subsection{Selecting Influential Examples}
\label{sec:secinf}
\begin{algorithm}[t!]
    \caption{SecInf($D_{trn}$,$D_{val}$,$r$) - Selecting Influential Examples}
    \begin{algorithmic}[1]
        \renewcommand{\algorithmicrequire}{ \textbf{Input:}}
        \renewcommand{\algorithmicensure}{ \textbf{Output:}}
        \Require Training data $D_{trn}$, validation data $D_{val}$, ratio of selected data $r$
        \Ensure $D_{s}$, $D_u$
        \State $D_{s} \leftarrow \emptyset$
        \For{$z_i \in D_{trn}$}
            \State Compute $\mathcal{I}_{up,loss}(z_i, D_{val})$ using Eq. \ref{eq:sg_up_loss}
        \EndFor
        \State Sort data points in $D_{trn}$ in descending order of absolute influence scores
        \State Assign the first $\lceil r\times |D_{trn}|\rceil$ examples of $D_{trn}$ to $D_{s}$
        \State $D_{u} \leftarrow D_{trn} \setminus D_{s}$ \\
        \Return $D_{s}$, $D_u$
    \end{algorithmic}
   
    \label{alg:selection}
\end{algorithm}

Recall from Section \ref{sec:prelims} that the influence function tells how the model loss would change if a data point $z$ was removed from the training set. In the presence of attacks, poisoned examples in training data can silently change the underlying data distribution. In other words, poisoning attacks shift the model's decision boundary away from the one over clean data, and the poisoned model effectively treats both poisoned and clean examples as normal training data. However, some poisoned examples are more effective than others in changing the model's predictions \cite{liu2022friendly,DBLP:conf/icml/YangLM22}. Intuitively, we can increase the generalization ability of the model by focusing on the training examples which have a higher impact. To do so, we introduce a subset selection method in Algorithm \ref{alg:selection}. For each example in the training data, we calculate the influence score using upweighting approach (Eq. \ref{eq:sg_up_loss}) as shown in lines 2-3. We then build the subset $D_s$ by selecting the $\lceil r\times |D_{trn}|\rceil$ examples which have the highest impact.

According to Basu et al.\cite{DBLP:conf/iclr/BasuPF21}, the influence function may not accurately estimate the change in loss when working with DNN models. The phenomenon is more noticeable when the models have deeper and wider architectures. The correctness of influence estimation also depends on various factors in the training scheme. However, even though the estimated changes in loss using Eq. \ref{eq:sg_up_loss} does not closely match the actual changes, they are still highly correlated \cite{DBLP:conf/icml/KohL17}. In fact, our Algorithm \ref{alg:selection} constructs the training subset $D_s$ based on the rank in the influence scores, not on the estimated change in loss.

\subsection{Adding Healthy Influential-Noise}
\label{sec:add_noise}
We first make an observation of how the influence score can help explain the effect of training input on the trained model.
Eq. \ref{eq:sg_pert_loss}, $\mathcal{I}_{pert, loss}(z, D_{val})$, tells us the contribution of each input pixel to the loss of the whole validation set. While the gradient of the loss of the validation set, $\nabla_{\theta}L(D_{val}, \hat{\theta})$, provides the information on how the trained model performs on the unseen validation set, the term $\nabla_x\nabla_{\theta}l(z, \hat{\theta})$ tells us how each input pixel contributes to the loss. Pixels with either significantly positive or negative influence scores will highlight the model's attention. Since the influence score represents the change in loss, a pixel that has a positive (negative) score will cause an increase (decrease) in the loss. Therefore, by crafting a noise in the opposite direction of influence score, i.e., $\delta=-\mathcal{I}_{pert, loss}(z, D_{val})$, we can perturb the original image in a way that strengthens pixels that have a helpful effect and weakens pixels that have a harmful effect. In the following example, we show that images that are predicted wrongly by the trained model can be altered by adding healthy noises based on their influence scores to increase the probability of being correctly classified by the model.
\begin{figure}[t!]
    \centering
    \includegraphics[width=.40\textwidth]{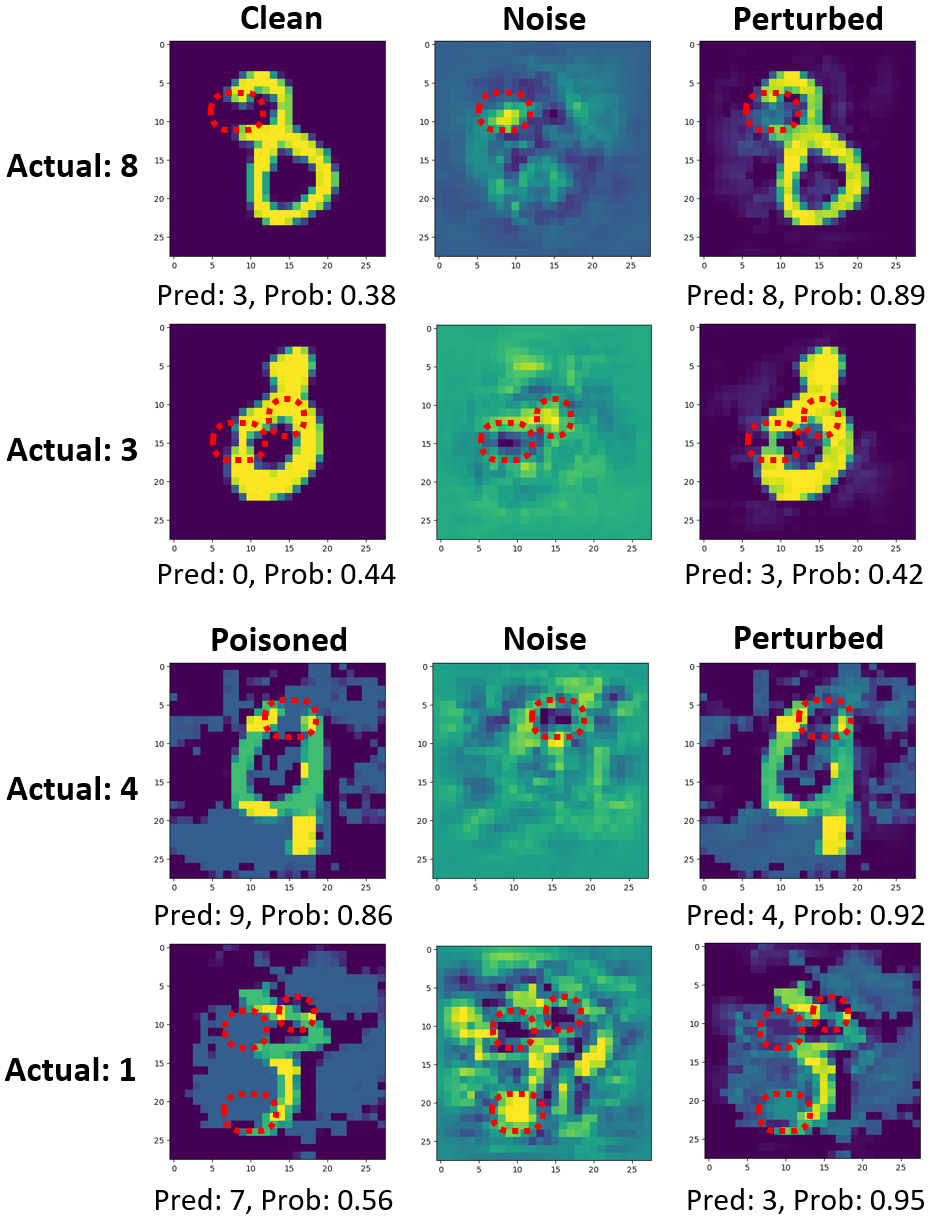}
    \caption{Using influence score to boost model prediction. The first two rows are clean examples. The last two rows are poisoned examples generated by DeepConfuse\cite{DBLP:conf/nips/FengCZ19}. Three columns from left to right are: original image, noise generated by HINT, and healthy-noise perturbed image. \textit{Pred} is the predicted class and \textit{Prob} is the probability. Red dotted circles are important regions that the influential noise focuses on.}
    \label{fig:motivation}
\end{figure}

In the second column of Figure \ref{fig:motivation}, we visualize the healthy noise for either clean or poisoned MNIST images. Note that the noise values can be either negative or positive, and we scale the values to be between 0 and 1 in the visualization. From the noise, we get distinct dark and bright regions which give information about the harmful/helpful regions inside the image. Each dark area corresponds to a harmful region, and these patches of pixels cause confusing regions that need to be reduced. On the other side, each bright area corresponds to a helpful region and depicts where healthy influential-noise can be added to improve the prediction ability of the model. Therefore, the influence function intuitively provides useful information on how we can perturb the original image to get better classification results from the model. We note that this conclusion is consistent with our observation from Eq. \ref{eq:sg_pert_loss}.

In the third column, we generate healthy-noise perturbed images by adding the healthy noise (middle column) to the original images (first column). By comparing the classification results of the third column with the results of the first column, we can see that adding noise based on the influence scores helps reduce the effect of harmful regions and boost the helpful regions of each image, as the correct class is now predicted with high probability. Therefore, the model benefits from the addition of healthy noise. Moreover, in the case of poisoned images, the healthy noise assigns negative values to the harmful poisoned regions so that the malicious perturbations become less effective. 

\begin{algorithm}[t!]
    \caption{AddNoise($\hat{D}_s$,$D_s$,$\gamma$,$\beta$) - Adding Healthy Influential-Noise}
    \begin{algorithmic}[1]
        \renewcommand{\algorithmicrequire}{ \textbf{Input:}}
        \renewcommand{\algorithmicensure}{ \textbf{Output:}}
        \Require Perturbed training subset $\hat{D}_s$ at previous update step, selected training subset $D_s$, scaling factor $\gamma$, healthy noise bound $\beta$
        \Ensure $\hat{D}_s$
        \For{$\hat{z}_i \in \hat{D}_{s}$ and $z_i \in D_s$}
            \State $\delta_i \leftarrow \hat{x}_i - x_i$
            \State $\delta_i \leftarrow \Pi_{\beta}(\delta_i - \gamma\mathcal{I}_{pert, loss}(\hat{z}_i, D_{val}))$
            \State $\hat{x}_i \leftarrow \text{Clip}(x_i+\delta_i)$
            \State $\hat{z}_i \leftarrow (\hat{x}_i, y_i)$
            \State Update new $\hat{z}_i$ in $\hat{D}_s$
        \EndFor
        \Return $\hat{D}_s$
    \end{algorithmic}
   
    \label{alg:noise_calc}
\end{algorithm}

Algorithm \ref{alg:noise_calc} demonstrates how our method generates healthy noise for each example in $D_s$. Let $\delta_i \in \Delta$ be the healthy influential-noise corresponding to training input $x_i$, i.e., $\hat{x}_i=x_i+\delta_i$ in line 2. We optimize the noise within $L_\infty$-norm $\beta$-bound. In other words, $\delta$ should belong to $\Delta = \{\delta \in \mathbb{R}^{H\times W}: \|\delta\|_{\infty} \leq \beta\}$. During an update round, for every training example $\hat{z}_i=(\hat{x}_i,y_i)\in \hat{D}_s$, we first generate the noise as $\mathcal{I}_{pert,loss}(\hat{z}_i,D_{val})$ (line 3) and then project it onto the feasible space. After that, we add the healthy noise to the training input and clip pixel values to be within a valid range (line 4). Finally, we update the newly perturbed examples on $\hat{D}_s$ in line 6.
We note that for $\mathcal{I}_{pert, loss}(\hat{z}_i, D_{val})$ in line 3, the differentiation involves all layers of the entire network to calculate $\nabla_x\nabla_{\theta}l\left(z, \hat{\theta}\right)$.


\subsection{Complexity Analysis}
In this section, we analyze the complexity of our proposed method. Let $p$ be the number of parameters in the model, $n$ be the size of the training set $D_{trn}$, $k$ be the size of the validation set $D_{val}$, and $d$ be the number of input features. In Algorithm \ref{alg:selection}, $\mathcal{I}_{up, loss}(z, D_{val})$ is computed for each training example before the inputs are sorted from most to least influential. Since $\left[\nabla_{\theta}\dfrac{1}{|D_{val}|}\sum_{i=1}^{\left|D_{val}\right|} l\left(z_{i}, \hat{\theta}\right)\right]^{\top}H_{\hat{\theta}}^{-1}$ is fixed, it only needs to be computed once -- which helps to reduce the overall running time. Algorithm \ref{alg:selection} requires $O(np)$ for calculating the loss of $n$ training examples and the one computation of IHVP using LiSSA takes $O(kp + rjp)$, where $r$ is the recursion depth and $j$ is the number of recursions. 
Sorting takes $O(n\log(n))$ on average, hence, the subset selection procedure takes in total $O(np + kp + rjp + n\log(n))$. Algorithm \ref{alg:noise_calc} generates and updates the healthy influential-noise for each example in the subset $\hat{D}_s$. The noise generation step in line 3 takes a similar running time to perform the IHVP estimation, and $O(dp)$ is the cost for calculating the gradient with respect to the input, $\nabla_x\nabla_{\theta}l\left(z, \hat{\theta}\right)$. Since $|D_s| \leq n$, Algorithm \ref{alg:noise_calc} takes $O(ndp + kp + rjp)$. In Algorithm \ref{alg:hint}, the training phase is performed over $T$ epochs, each of which needs $O(np)$ to update model's parameters. The training pipeline calls SecInf (Algorithm \ref{alg:selection}) once in line 4, and AddNoise  (Algorithm \ref{alg:noise_calc}) $s$ times in line 6-9, where $s=|S|$. Since $k\ll n$ and $s \ll T$, the total complexity for our HINT (Algorithm \ref{alg:hint}) is $O(Tp(nd + rj) + n\log(n))$.



\section{Data Poisoning Attacks and Defenses}
\label{sec:atk_def}
\subsection{Data Poisoning Attacks}
\label{sec:poison_attacks}
 We detail the untargeted and targeted attacks we utilize in our experiments to craft adversarial examples.

\subsubsection{Untargeted Attacks} The untargeted attacks we consider include projected gradient descent (PGD) \cite{DBLP:conf/iclr/MadryMSTV18}, delusive adversarial perturbation (DAP) \cite{DBLP:conf/nips/TaoFYHC21}, delusive universal random perturbation (DURP) \cite{DBLP:conf/nips/TaoFYHC21}, and deep confuse (DC) \cite{DBLP:conf/nips/FengCZ19}. PGD was originally proposed as a test-time attack and it utilizes gradient information to generate adversarial perturbations. DAP crafts an adversarial training input $\Tilde{x}_i$ by minimizing the loss $l(f_\theta(\Tilde{x}_i), t_i)$ where $t_i$ is a class other than $y_i$ and $\Tilde{x}$ is bounded by a small tolerance rate $\epsilon$. DURP works by adding class-wise random perturbation $\mu(y_i)$ to each $x_i$. This means that for all training examples having class $y_i$, the attack will add the same perturbation $\mu(y_i)$ to them. DC generates malicious examples using an Auto-Encoder architecture (e.g., UNet for the CIFAR-10 dataset). This attack can craft imperceptible and efficient poisoning examples.

\subsubsection{Targeted Attacks} The attacker's objective can be formulated as a bi-level optimization problem:
\begin{equation}
    \min_{\epsilon \in C} l(f_{\theta_\epsilon}(x_t), y_{adv}) \hspace{0.5em} \text{s.t.} \hspace{0.5em} \theta_\epsilon = \argmin_{\theta \in \Theta}l(f_{\theta}(x_i+\epsilon_i),y_i) 
    \label{eq:tgt_bilevel}
\end{equation}
where $C=\{\epsilon \in \mathbb{R}^{H\times W}:||\epsilon||_\infty \leq \xi, \epsilon_i=0 \ \forall i \notin D_p \}$ is the constraint set of malicious perturbation $\epsilon$, and $D_p$ is the poisoned set.
Commonly, the malicious perturbations lie within $\xi$-bounded $l_\infty$ ball to be imperceptible.
We consider four different targeted attacks in our experimental evaluation: MetaPoison (MP) \cite{DBLP:conf/nips/HuangGFTG20}, gradient matching (GM) \cite{DBLP:conf/iclr/GeipingFHCT0G21}, bullseye polytope (BP) \cite{DBLP:conf/eurosp/AghakhaniMWKV21}, and feature collision (FC) \cite{DBLP:conf/nips/ShafahiHNSSDG18}. In four attacks, GM and MP are two modern attacks in training-from-scratch scenario, while BP and FC work efficiently in transfer learning scenario. 

MP uses a meta-learning approach to approximate the attacker's bilevel objective in Eq. \ref{eq:tgt_bilevel}. To do so, MP runs multiple unroll steps to approximate the inner optimization, and looks into the training pipeline to evaluate how the perturbation will affect the adversarial loss in future training steps. Then, the method uses the Adam optimizer to update the perturbation. 
GM optimizes the malicious perturbation via aligning the gradients of targeted and poisoned examples. Therefore, the method only needs one unroll step to compute the gradient. To optimize the malicious perturbation, GM attempts to minimize the negative cosine similarity between adversarial loss and natural loss, where the adversarial (natural) loss is the loss defined in the outer (inner) objective function in Eq. \ref{eq:tgt_bilevel}. In BP, the attacker crafts poisoned examples such that their representation in feature space is close to the targeted image. BP significantly improves the scalability and transferability of Convex Polytope attack\cite{DBLP:conf/icml/ZhuHLTSG19}. FC, also known as Poison Frogs attack, has a similar idea to BP that it explores the feature space of images. The method aims to optimize the malicious examples such that they collide with the targeted example in the feacture space.  

\subsection{Defenses against Data Poisoning Attacks}
\label{sec:dadp} 
In this section, we detail the defense mechanisms that we compare with HINT in Section \ref{sec:experiments}. Specifically, we consider the FRIENDS\cite{liu2022friendly}, adversarial training against delusive adversaries (ATDA)\cite{DBLP:conf/nips/TaoFYHC21}, and EPIC\cite{DBLP:conf/icml/YangLM22} algorithms.

\noindent\textbf{FRIENDS.} From the observation that each effective poison causes a local increase in the training loss and the whole poisoning set forms local regions in the loss space, FRIENDS aims to optimize the maximum perturbation without changing the model prediction. For each training example $x_i$, the friendly noise $\epsilon_i$ is generated as:
\begin{equation*}
    \epsilon_i = \argmin_{\epsilon:||\epsilon||_{\infty}\leq \beta}D_{KL}(f_\theta(x_i+\epsilon)||f_\theta(x_i))-\lambda ||\epsilon||_2,
\end{equation*}
where $\lambda$ is the scaling factor. Besides the friendly noise, FRIENDS adds random noise from Gaussian, Uniform, or Bernoulli distribution to smooth the training loss. 

\noindent\textbf{ATDA.} Following the theoretical proof that the adversarial risk can be the upper bound of natural risk, ATDA adapts the adversarial training technique to defend against untargeted poisoning attacks. By using FAT\cite{DBLP:conf/icml/ZhangXH0CSK20}, the adversarial example $\Tilde{x}_i$ can be generated as:
\begin{equation*}
    \begin{split}
        \Tilde{x}_i = & \argmin_{\Tilde{x}:||\Tilde{x}-x||_{p}\leq \beta}  l(f_\theta(\Tilde{x}_i),y_i) \\
        \text{s.t.}\  & l(f_\theta(\Tilde{x}_i),y_i)-\min_{y\in Y}l(f_\theta(\Tilde{x}_i),y) \geq \tau,
    \end{split}
\end{equation*}
where $\tau > 0$ is the margin such that an adversarial example would be misclassified.

\noindent\textbf{EPIC.} Different from the previously discussed methods, EPIC finds and drops malicious training points. From the observation that effective poisoned examples are often isolated from others of the same class in the gradient space, EPIC builds a set of medoids of each class, assigns other data points to its closest medoid, and drops isolated medoids during the training. The objective to find the set of medoids can be formulated as:
\begin{equation*}
    S \in \argmin_{S \subseteq V, |S|<m} \sum_{i\in V}\min_{j\in S}||\nabla l(f_\theta(x_i),y_i)-\nabla l(f_\theta(x_j),y_j)||_2,
\end{equation*}
where $m$ is the maximum number of medoids, $S$ is the index set of medoids, and $V$ is the index set of training data.

\noindent\textbf{Remark.} Our HINT is similar to FRIENDS and ATDA in principle as all three methods perturb training examples to defend against poisoning attacks. But HINT uses a different defense mechanism as aforementioned in Section \ref{sec:hint}. The healthy noise can capture local harmful/helpful regions formed by influential pixels and, by focusing on those important regions, the trained model is more resilient to attacks. Furthermore, HINT leverages the advantage of influence function, which can estimate the impact of each training example to the model loss, to choose a subset of training examples for perturbation.

\section{Experiments}
\label{sec:experiments}
\subsection{Evaluation Setup}

\subsubsection{Dataset and Model}
\label{sec:dataset}
We focus on the image classification tasks and use MNIST and CIFAR-10 as our primary evaluation datasets. Each dataset is divided into three subsets (train, validation, and test), and a classification model is trained over the training set. Table \ref{tab:data_model} gives the specific details of our dataset construction. For MNIST, we train the CNN model using the SGD method with a learning rate of 0.01. For CIFAR-10, we train the ResNet-18 model using the SGD method with a Nesterov momentum of 0.9 and weight decay of $5\times 10^{-4}$. Data augmentation techniques such as random crop and horizontal flip are also applied to the training images. Additionally, the initial learning rate is set to 0.1, which is then decreased by a factor of 10 at epochs 30, 50, and 70. 
\begin{table}[ht]
    \centering
    \caption{Description of datasets and corresponding models.}
    \label{tab:data_model}
    \begin{tabular}{c|c|c|c|c|c|c}
        \hline
        Dataset & $|D_{trn}|$ & $|D_{val}|$ & $|D_{tst}|$ & Model & Batch & Epoch \\
        \hline
        MNIST & 59000 & 1000 & 10000 & CNN & 128 & 30 \\
        CIFAR-10 & 49000 & 1000 & 10000 & Resnet-18 & 128 & 80 \\
        \hline
    \end{tabular}
\end{table}

\subsubsection{Poisoning Training Data}
We use the attack algorithms aforementioned in Section \ref{sec:poison_attacks} to generate adversarial examples and inject them into the clean training data to produce our poisoned training set. 

\noindent\textbf{Poisoning training data with untargeted attacks.}
In this experiment, we consider the difficult scenario where the poisoned training data contains poisoned examples generated by four different malicious attacks: PGD, DAP, DURP, and DC. We train a victim model over the clean data for each attack type and generate the same amount of poisoned data per attack type based on a poison ratio $\rho$. For example, using CIFAR-10 with $\rho=0.4$, we create a poisoned training dataset (with 49,000 images), which has 29,400 clean images and 19,600 poisoned images (4,900 images per attack type). For each attack type, the attacker's budget for perturbation is $\xi=0.031$ (or 8/255) when attacking CIFAR-10 and $\xi=0.3$ (or 76/255) when attacking MNIST. The setting is practical since the attacker tries to use a combination of multiple powerful attacks to efficiently generate malicious data.  

\noindent\textbf{Poisoning training data with targeted attacks.}
When analyzing the ability of defenses against targeted attacks, we only consider the CIFAR-10 as this dataset is commonly used for targeted attack evaluation. In this experiment, we evaluate the effectiveness of defense methods in both training-from-scratch (with GM and MP) and transfer learning (with BP and FC) settings as aforementioned in Section \ref{sec:poison_attacks}. In both scenarios, we assume that the attacker knows the dataset, model architecture, and training scheme. However, they do not know the model weights.
Similar to other papers aimed at defending against targeted attacks \cite{DBLP:conf/icml/YangLM22, liu2022friendly}, and the benchmarks in \cite{DBLP:conf/icml/SchwarzschildGG21}, we randomly choose the target and source classes, and then generate 490 poisoned training images ($\rho = 1\%$). 

We run the attacks with the same setup, except for some specific hyper-parameters for each attack that we mention in this section. For GM, BP and FC, we train the victim model for 80 epochs and choose $\xi=0.062$ (or 16/255) as the bound for malicious perturbation. The number of attack iterations is set at 250, 4,000, and 1,000, respectively. Other hyper-parameters follow the default values from the implementation of GM\footnote{\href{https://github.com/JonasGeiping/poisoning-gradient-matching}{https://github.com/JonasGeiping/poisoning-gradient-matching}}. For MP, the poisoned data is downloaded from MetaPoison\footnote{\href{https://github.com/wronnyhuang/metapoison}{https://github.com/wronnyhuang/metapoison}}. We use an equal mix of poison-dog target-bird and poison-frog target-plane settings in the evaluation. The bound for perturbation used in MetaPoison is $\xi=0.031$. 

For transfer learning scenario, we use a similar setup as in the from-scratch scenario to pre-train the victim model. Then, we randomly re-initialize the top layers while freezing the feature extraction layers. The victim model is then optimized on the transfer set. We construct the clean training set (for pre-training the victim model) with 44,100 clean images (90\%) selected uniformly from each class. The transfer set consists of the remaining 4,900 images (10\%) of the training set, in which 490 examples have been poisoned. 
We note that transfer learning setting used in our evaluation is not a real transfer learning setting as the clean training and the transfer sets come from the same original dataset. This, however, is the worst-case scenario that a defense has to consider to show its effectiveness, as the setting makes it easier for the attacker to succeed \cite{DBLP:conf/iclr/GeipingFHCT0G21,liu2022friendly}.

\subsubsection{HINT and Defense Baselines} 
In this section, we briefly discuss the hyper-parameters for running each defense used in our experiment. 

\noindent\textbf{Defense Baselines.} We choose FRIENDS+Bernoulli in our evaluation as it has the best performance according to \cite{liu2022friendly}. Similar to the default setting of FRIENDS, we set $\beta=0.062$. For ATDA, we choose FAT as the representation and set the default defender's budget $\beta$ as 0.25 based on the findings presented in \cite{DBLP:conf/nips/TaoFYHC21}. When conducting experiments of ATDA against targeted attacks, we also use $\beta=0.062$ for a fair comparison with other baselines. 
EPIC-0.1 is the default representation for EPIC with the poison drop interval $T$ and warm-up epochs $K$ as: $T=4$, $K=10$ for CIFAR-10; and $T=2$, $K=5$ for MNIST. We note that our evaluation uses default values as presented in the FRIENDS, ATDA, and EPIC papers \cite{liu2022friendly,DBLP:conf/nips/TaoFYHC21,DBLP:conf/icml/YangLM22} for all other hyper-parameters. Besides the above defenses, we additionally run experiments for a na\"{i}ve training method (W/o Defense) using the default architecture of each dataset.


\noindent\textbf{HINT.} When computing the healthy noise, we only use the weights from the top layers of the model and discard all other weights (i.e., all weights from feature extraction layers). We set the update schedule for healthy noise at epochs 5, 15, and 40 for CIFAR-10 and at epochs 5 and 15 for MNIST. We choose the budget for the healthy noise to be $\beta=0.062$ and the scaling factor to be $\gamma=0.1$. 

\noindent\textbf{Metrics.} 
For each experiment, we report the average and standard deviation of the test accuracy over five trials. When defending against targeted attacks, we additionally report the Attack Success Rate (ASR), which evaluates the success of an attack in changing the prediction of targeted examples. Note that an attack is considered successful only if it can change the predicted class to the intended class, following the evaluation in \cite{DBLP:conf/iclr/GeipingFHCT0G21,liu2022friendly,DBLP:conf/icml/YangLM22}. We use GPU Tesla V100 (32GB RAM) and CPU Xeon 6258R 2.7 GHz to conduct all experiments.


\subsection{Results}
\subsubsection{Defending against Untargeted Attacks} 
In this experiment, we evaluate the ability of different defense mechanisms to defend against poisoning attacks under different poison ratios. In Table \ref{tab:multi_attks}, we report the average and standard deviation of the test accuracy on CIFAR-10 and MNIST datasets. Each row shows the result for one poisoning set constructed from a particular poison ratio $\rho$. In most rows, HINT outperforms all other defense baselines, demonstrating our defense's effectiveness against poisoning attacks. For HINT and other methods, the test accuracy decreases when $\rho$ increases, which is not surprising since the attacker is able to inject more poisoning examples into the training set. With a large poison ratio ($p\geq 0.8$), our HINT method significantly reduces the effect of poisoned samples when we compared to na\"{i}ve training. It also clearly outperforms other three defense baselines. 

When we look at the results on the CIFAR-10 with a small poison ratio ($\rho \leq 0.4$), HINT is the only defense method that achieves better performance than na\"{i}ve training, while other defenses have significant gaps below (more than $1.5\%$). A similar pattern happens to the results on the MNIST when $\rho \leq 0.6$. In this case, the model still learns both clean and malicious patterns but is able to focus more on the clean data. With FRIENDS and ATDA, when friendly or adversarial noises are added to clean training examples, the decision boundary may move further away from the original one. Recall from Section \ref{sec:dadp} where we explain that FRIENDS and ATDA add noise to the whole training dataset while our HINT method perturbs only a selected subset. Especially in the case that there is no poisoning attack performed ($\rho=0.0$), our HINT method only loses $0.22\%$ test accuracy compared to na\"{i}ve training on MNIST and even has better accuracy on CIFAR-10. These results match our previous observation in Section \ref{sec:add_noise} that the model trained with healthy noise does not lose its generalization ability. 
On both datasets, EPIC is the worst performer in the four defenses. This is because EPIC continuously detects and drops malicious examples, which works more efficiently when poisoned examples are far from their class in the gradient space. However, it is hard for EPIC to detect poisoned examples by untargeted attacks since untargeted attacks significantly perturb multiple training examples of every class to shatter the decision boundary, which also tampers the representations of training examples in the gradient space. It is even more challenging when we mix four different attack types with clean data in our setting. 

\begin{figure}[t!]
    \centering
    \includegraphics[width=.49\textwidth]{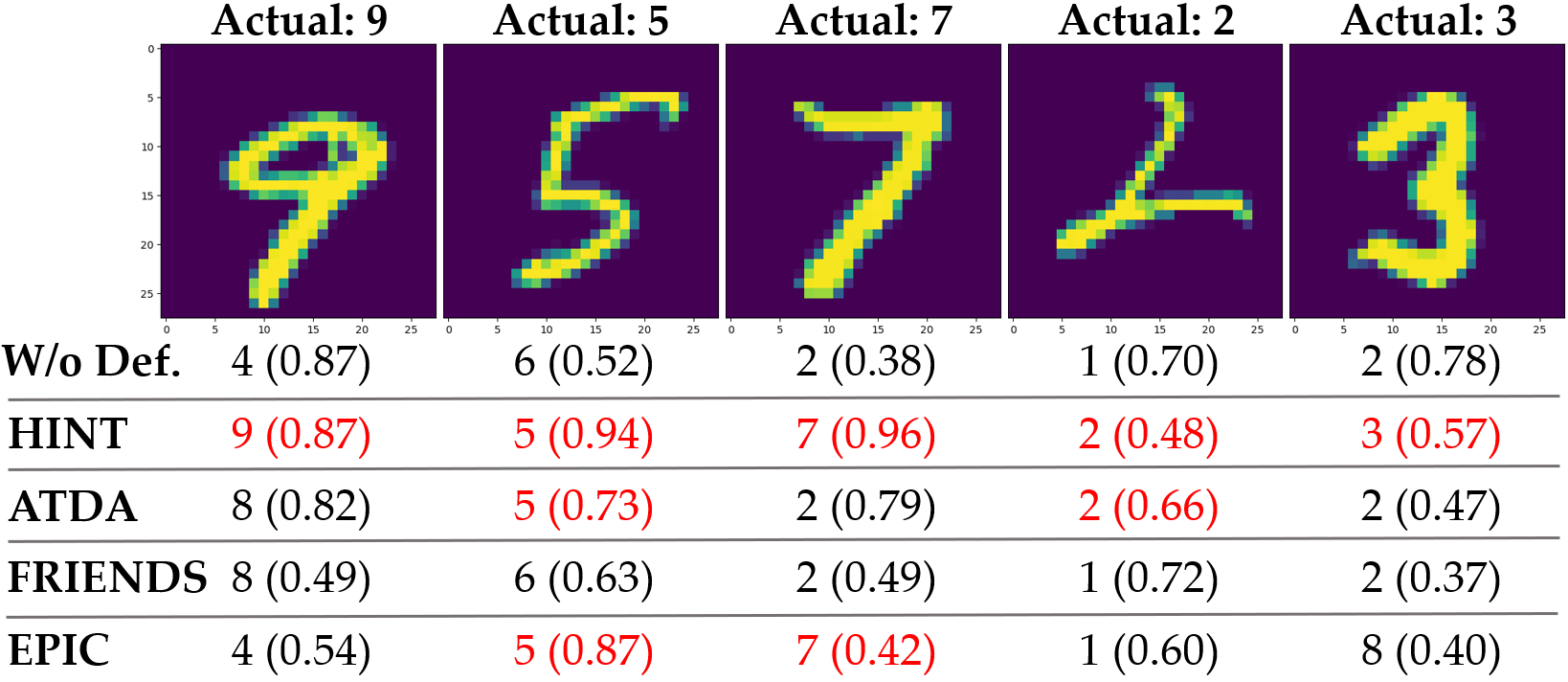}
    \caption{Prediction of defense methods on MNIST test images under multiple untargeted attacks.}
    \label{fig:mnist_test_defense}
\end{figure}

In Figure \ref{fig:mnist_test_defense}, we show the predicted class, along with the probability of the prediction, of HINT and other baselines for five test examples from MNIST. We use red to denote each defense's success in preventing untargeted attacks. Each result shows the predicted class with the confidence score from the trained model using the corresponding defense method. Some pairs of classes have a high chance of confusion in prediction, which shows that untargeted attacks can successfully mislead the victim model and shift the decision boundary. For example, a trained model easily gets confusing an image of digit 7 for digit 2, or an image of digit 5 for digit 6. The results show that our method is more effective than other baselines in helping the trained model avoid the effect of untargeted attacks and give correct prediction results with a high confidence score.

\begin{table*}[ht!]
    \centering
    \caption{Test accuracy (\%) of our method and baselines when defending against multiple untargeted attacks (PGD+DAP+DURP+DC) on CIFAR-10 and MNIST datasets. In this evaluation, $r=0.5$ and $\rho$ is the poison ratio.}
    \label{tab:multi_attks}
    \begin{tabular}{c|c|c|c|c|c|c}
        \hline
        Dataset & $\rho$ & W/o Defense & HINT & ATDA & FRIENDS & EPIC\\ \hline
        \multirow{6}{*}{CIFAR-10} & 0.0 & \textbf{93.86 $\pm$ 0.26} & 93.64 $\pm$ 0.12 & 91.82 $\pm$ 0.17 & 89.25 $\pm$ 0.53 & 87.28 $\pm$ 0.71 \\
        &0.2 & 92.54 $\pm$ 0.05 & \textbf{92.67 $\pm$ 0.04} & 90.97 $\pm$ 0.17 & 89.37 $\pm$ 0.52 & 86.98 $\pm$ 0.79 \\
        &0.4 & 91.67 $\pm$ 0.12 & \textbf{92.08 $\pm$ 0.17} & 89.92 $\pm$ 0.25 & 88.53 $\pm$ 0.38 & 87.37 $\pm$ 0.40 \\
        &0.6 & 83.51 $\pm$ 0.17 & \textbf{91.94 $\pm$ 0.21} & 90.00 $\pm$ 0.38 & 88.90 $\pm$ 0.53 & 86.70 $\pm$ 0.11 \\
        &0.8 & 80.02 $\pm$ 0.36 &  \textbf{91.20 $\pm$ 0.21} & 89.22 $\pm$ 0.57 & 87.75 $\pm$ 0.18 & 85.38 $\pm$ 0.73\\
        &1.0 & 48.62 $\pm$ 0.74  & \textbf{90.79 $\pm$ 0.13} & 89.41 $\pm$ 0.58 & 86.76 $\pm$ 0.55 & 84.73 $\pm$ 0.32 \\
        \hline
        \multirow{6}{*}{MNIST} & 0.0 & 98.44 $\pm$ 0.02 & \textbf{98.87 $\pm$ 0.01} & 98.45 $\pm$ 0.11 & 98.33 $\pm$ 0.09 & 98.06 $\pm$ 0.05\\
        &0.2 & 97.73 $\pm$ 0.01 & \textbf{98.42 $\pm$ 0.02} & 97.39 $\pm$ 0.29 & 97.58 $\pm$ 0.08 & 97.31 $\pm$ 0.16 \\
        &0.4 & 97.15 $\pm$ 0.03 & \textbf{98.01 $\pm$ 0.09} & 97.01 $\pm$ 0.09 & 97.14 $\pm$ 0.12 & 96.76 $\pm$ 0.19 \\
        &0.6 & 96.64 $\pm$ 0.12 & \textbf{96.87 $\pm$ 0.02} & 96.46 $\pm$ 0.14 & 96.58 $\pm$ 0.09 & 96.19 $\pm$ 0.19 \\
        &0.8 & 95.55 $\pm$ 0.12 & \textbf{95.88 $\pm$ 0.04} & 95.59 $\pm$ 0.09 & 95.49 $\pm$ 0.21 & 95.03 $\pm$ 0.20 \\
        &1.0 & 74.62 $\pm$ 0.94 & \textbf{88.92 $\pm$ 0.19} & 80.60 $\pm$ 1.01 & 77.08 $\pm$ 1.70 & 77.84 $\pm$ 1.54 \\
        \hline
    \end{tabular}
\end{table*}

\subsubsection{Defending against Targeted Attacks}
Table \ref{tab:tgt_atks} shows a comparison of our HINT method with all baselines in terms of ASR and test accuracy on CIFAR-10. We compare the results based on ASR (lower is better) and test accuracy (higher is better). In this setting, our HINT method also significantly outperforms all other baselines. With all attacks, the na\"{i}ve model fails all five trials, except for MP with 4/5. In particular, HINT, FRIENDS, and EPIC succeed in keeping the targeted example safe under the GM attack. However, our method achieves the highest test accuracy of the three defenses. Under other attack methods, HINT only fails once, while most other baselines fail multiple times. In most experiments, our HINT method also consistently has the best test accuracy compared to the other baselines. ATDA is the worst performer in terms of ASR compared to HINT, FRIENDS, and EPIC. This is unsurprising since ATDA uses adversarial perturbation to break the malicious examples by untargeted attacks, while the training data is poisoned by targeted attacks in this experiment. 

From the results, we can see that all the defense methods work well when defending against GM and MP. In training-from-scratch scenario, when training the model from scratch, defense methods will have a better chance to capture the malicious pattern from the poisoned examples and thereby be able to reduce their effect. Conversely, FRIENDS and EPIC have noticeably dropped in efficiency when defending against BP and FC in transfer learning scenario. 

\begin{figure}[ht]
    \centering
    \includegraphics[width=.49\textwidth]{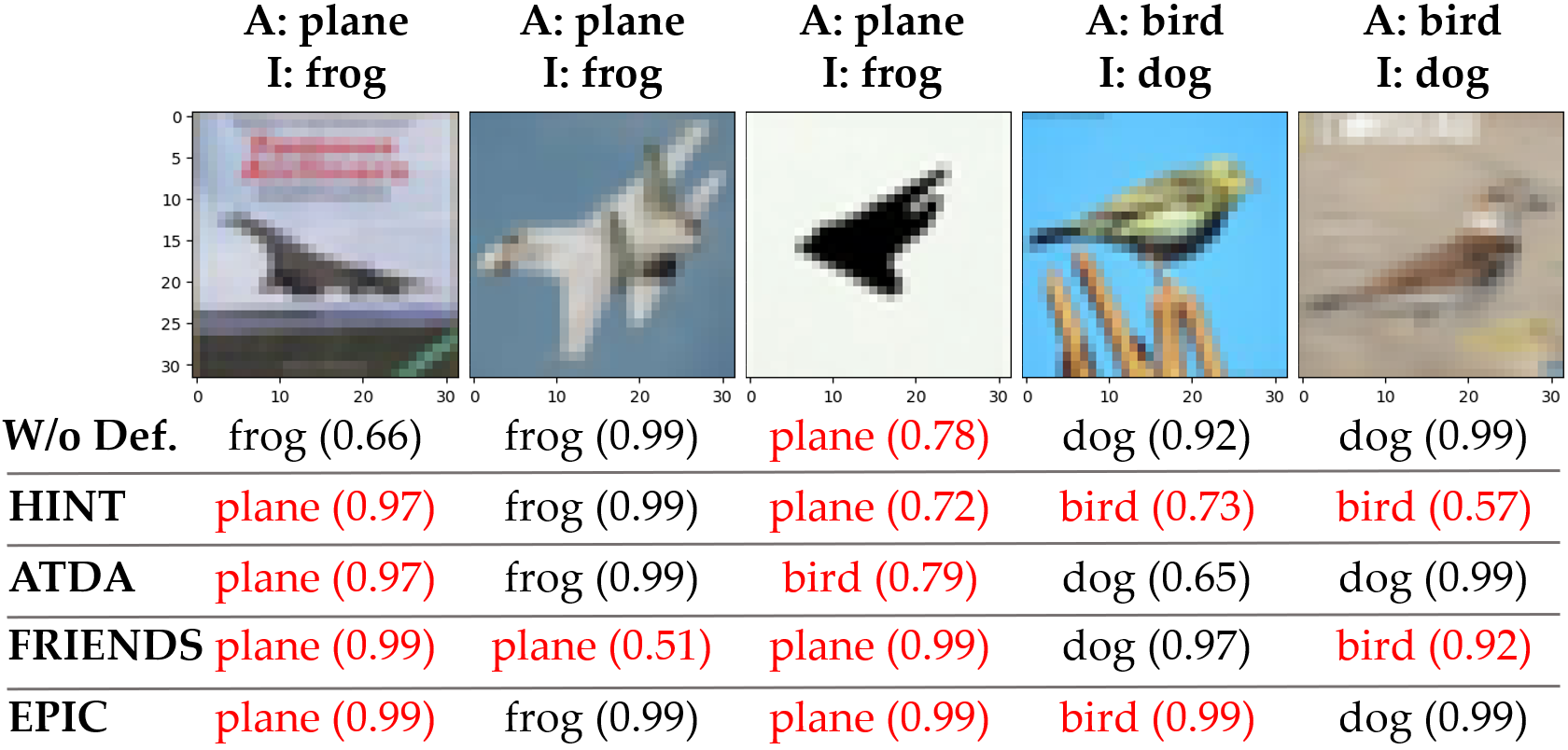}
    \caption{Prediction of defense methods on CIFAR-10 test images under MP attack. ``A" and``I" stand for actual and intended classes, respectively.}
    \label{fig:cifar10_mp_test_defense}
\end{figure}

\begin{figure}[ht]
    \centering
    \includegraphics[width=.49\textwidth]{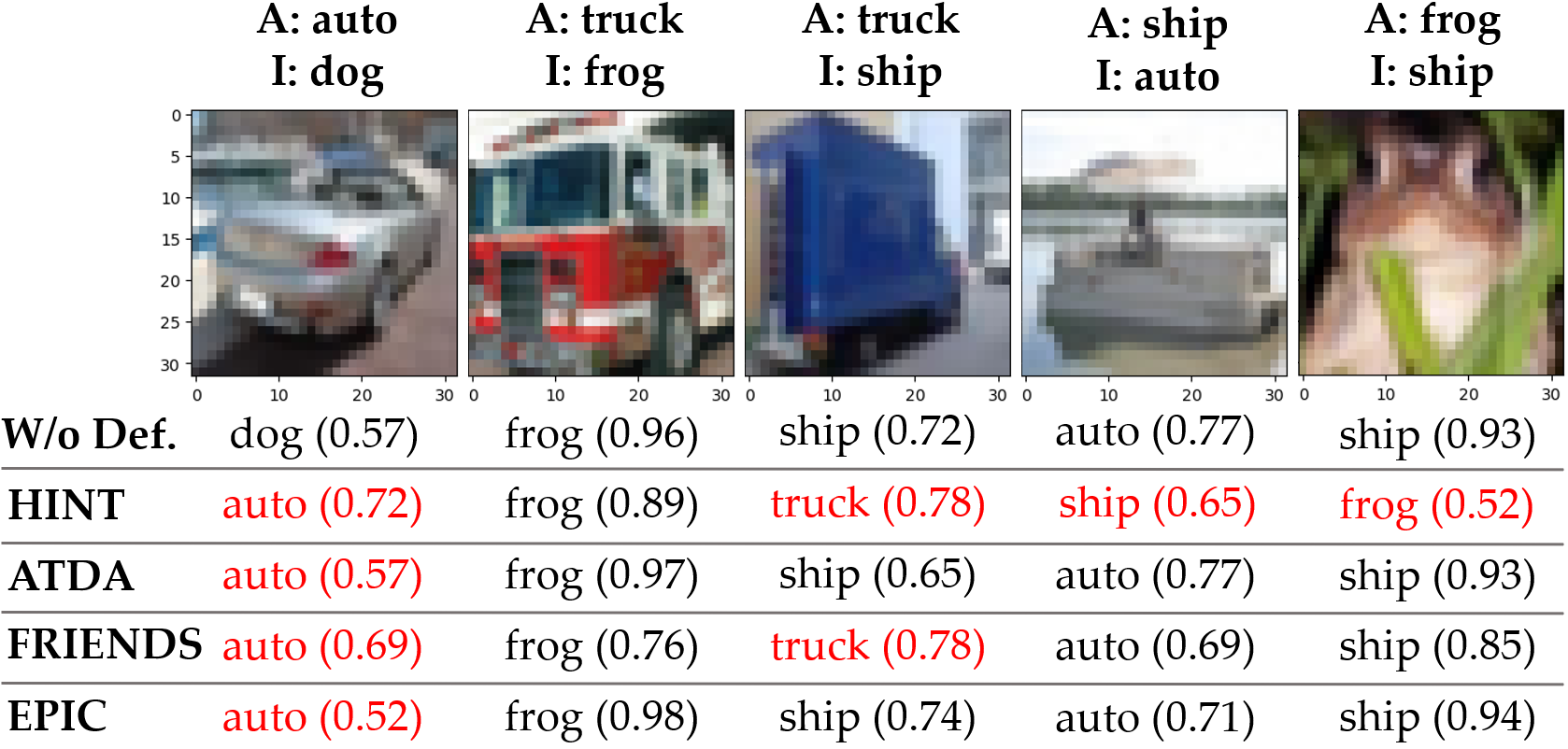}
    \caption{Prediction of defense methods on CIFAR-10 test images under BP attack. ``A" and``I" stand for actual and intended classes, respectively.}
    \label{fig:cifar10_bp_test_defense}
\end{figure}

Figure \ref{fig:cifar10_mp_test_defense} and \ref{fig:cifar10_bp_test_defense} illustrate our results under MP and BP attacks, respectively. Similar to Figure \ref{fig:mnist_test_defense}, we use red color to denote the defense's success in preventing attacks from falsifying the target's class, and each result shows the predicted class with the confidence score. The model trained with HINT predicts the correct classes for most of the targeted examples with a high confidence score, consistent with the results of defending against untargeted attacks. Especially under the BP attack, HINT is the only method to successfully defend against the attack trials in the last two columns.

\begin{table*}[ht]
    \centering
    \caption{Attack Success Rate and test accuracy (\%) of defense mechanisms against different targeted poisoning attacks. Experiments with MetaPoison is run without augmentation, following the setting in \cite{DBLP:conf/nips/HuangGFTG20,liu2022friendly}.}
    \label{tab:tgt_atks}
    \begin{tabular}{c|c|c|c|c|c}
        \hline
         & & GM & MP & BP & FC \\
        \hline
        \multirow{2}{*}{W/o Defense} & ASR & 5/5 & 4/5 & 5/5 & 5/5 \\
        & Test acc. & 93.69 $\pm$ 0.19 & 87.48 $\pm$ 0.41 & 91.41 $\pm$ 1.34 & 89.40 $\pm$ 1.39  \\
        \hline
        \multirow{2}{*}{HINT} & ASR & \textbf{0/5} & \textbf{1/5} & \textbf{1/5} & \textbf{1/5} \\
        & Test acc. & \textbf{92.99 $\pm$ 0.26} & \textbf{87.15 $\pm$ 0.46} & \textbf{92.41 $\pm$ 0.54} & \textbf{92.21 $\pm$ 0.31} \\
        \hline
        \multirow{2}{*}{ATDA} & ASR & 3/5 & 3/5 & 4/5 & 4/5 \\
        & Test acc. & 93.64 $\pm$ 0.27 & 87.45 $\pm$ 0.61 & 89.50 $\pm$ 1.51 & 88.37 $\pm$ 1.73 \\
        \hline
        \multirow{2}{*}{FRIENDS} & ASR &  0/5  & 1/5 & 3/5 & 2/5 \\
        & Test acc. & 89.17 $\pm$ 0.41 & 78.26 $\pm$ 0.63 & 89.53 $\pm$ 0.66 & 88.84 $\pm$ 0.94 \\
        \hline
        \multirow{2}{*}{EPIC} & ASR & 0/5 & 2/5 & 4/5 & 3/5 \\
        & Test acc. & 90.36 $\pm$ 0.43 & 86.68 $\pm$ 0.23 & 89.65 $\pm$ 2.34 & 89.19 $\pm$ 1.61 \\
        \hline
    \end{tabular}
\end{table*}

\subsubsection{Sensitivity Analysis}
Table \ref{tab:sent_analysis} shows the sensitivity analysis of HINT on the ratio of selected examples $r$ and the defender's budget $\beta$. We run our HINT method with different $r$ values to show how the size of the selected subset affects the model's performance. 
Note that when $r=1.0$, the result is equivalent to the case of removing the subset selection module. The result shows that as $r$ increases, the test accuracy decreases. This observed trend highlights the contribution of our subset selection module in the whole training method. The trend also matches our previous observation that adding noise into clean examples may cause a drop in test accuracy. 

For hyper-parameter $\beta$, the test accuracy increases when moving from $0.031$ to $0.062$, and decreases when $\beta > 0.062$. This phenomenon is understandable since the healthy noise is not large enough to reduce the effect of malicious perturbation when we choose small $\beta$. However, when the healthy noise is sufficiently large, it makes the values of the pixels move too far from their original value and breaks the spatial relationship in the images.
\begin{table}[ht]
    \centering
    \caption{Sensitivity analysis of HINT on $r$ and $\beta$ with untargeted attacks ($\rho=0.6$).}
        \begin{tabular}{c|c|c|c}
        \hline
          & Value  & CIFAR-10 & MNIST\\ \hline
         \multirow{4}{*}{$r$} & 0.25 & 92.06 $\pm$ 0.08 & 97.13 $\pm$ 0.03 \\
         & 0.5 & 91.94 $\pm$ 0.21 & 96.87 $\pm$ 0.02  \\
         &0.75 & 91.83 $\pm$ 0.11 & 96.62 $\pm$ 0.03 \\
         &1.0 & 91.72 $\pm$ 0.15 & 96.46 $\pm$ 0.05\\
        \hline
        \multirow{4}{*}{$\beta$} & 0.031 & 91.80 $\pm$ 0.18 & 96.76 $\pm$ 0.01  \\
         & 0.062 & 91.94 $\pm$ 0.21  & 96.87 $\pm$ 0.02 \\
         & 0.125 & 91.77 $\pm$ 0.13  & 96.74 $\pm$ 0.05 \\
         & 0.251 & 91.40 $\pm$ 0.20  & 96.43 $\pm$ 0.01 \\
         \hline
        \end{tabular}
    \label{tab:sent_analysis}
\end{table}

\subsubsection{Execution Time}
We report the execution time of all defense methods used in our experiments in Table \ref{tab:time}.  For HINT, we report the time for four variants with different $r$ values. When $r$ increases, the running time increases since the method needs to compute healthy noise for more training examples. 
When $r=1.0$, the running time does not significantly increase compared to when $r=0.75$ since HINT disables the subset selection module. 
For other defenses, the running time of FRIENDS is close to the default setting of HINT ($r=0.5$), while ATDA and EPIC need more time to execute. ATDA generates and adds the noise for every training example in every training step. EPIC executes poison identification, which requires extensive resources to find medoids for each class in gradient space, to drop isolated points in every $T$ epoch. 
\begin{table}[ht]
    \centering
    \caption{Execution time of defenses with CIFAR-10.}
    \begin{subtable}[h]{0.24\textwidth}
        \centering
        \begin{tabular}{c|c}
                \hline
                Method  & Time (s) \\ \hline
                HINT ($r=0.25$)  & 2113  \\
                HINT ($r=0.5$)  & 2427  \\
                HINT ($r=0.75$)  & 2745  \\
                HINT ($r=1.0$)  & 2875 \\
                \hline
        \end{tabular}
    \end{subtable}
    \hfill
    \begin{subtable}[h]{0.24\textwidth}
        \centering
        \begin{tabular}{c|c}
                \hline
                Method  & Time (s) \\ \hline
                W/o Defense & 1600 \\
                ATDA  & 5760 \\
                FRIENDS  & 2426 \\
                EPIC & 3147 \\
                \hline
        \end{tabular}
    \end{subtable}
    \label{tab:time}
\end{table}

\section{Conclusion}
\label{sec:conclusion}
In this work, we presented an effective robust training framework, HINT, that hardens the model with healthy influential-noise to protect machine learning models from poisoning attacks. Our method uses the influence function as a central mechanism to select examples with the highest impact on the model test loss and crafts the healthy influential-noise. 
Deep learning models trained with our HINT method are more resilient to the effect of malicious examples. Through comprehensive empirical evaluations, we demonstrate the effectiveness and stability of HINT in defending against powerful untargeted and targeted attacks (e.g., Deep Confuse, Gradient Matching, and Bulleyes Polytope) and its superiority over state-of-the-art defense baselines. These evaluations were conducted in a realistic scenario, highlighting the suitability of our defense mechanism for deployment in sensitive security settings. In future work, we will extend our approach to defend against other attack types, such as back door attacks, which involve injecting specific backdoor patterns into selected training data, manipulating test data to embed the triggers, and causing intentional misclassification.  

\noindent\textbf{Reproducibility.} Our source code is available at \url{https://github.com/minhhao97vn/HINT}.

\section*{Acknowledgements}
This work was supported  in part by National Science Foundation under awards 1946391, the National Institute of General Medical Sciences of National Institutes of Health under award P20GM139768, and the Arkansas Integrative Metabolic Research Center at University of Arkansas.

\bibliographystyle{IEEEtran}
\bibliography{IEEEabrv,main.bbl}

\end{document}